\documentclass{article}
\usepackage{spconf, amsmath, amsfonts, graphicx, multirow}
\begin{document}

\title{PointHop++: A Lightweight Learning Model on Point Sets \\ for 3D Classification
\thanks{This work was supported by Tencent.}}

\name{Min Zhang$^{1}$, Yifan Wang$^{1}$, Pranav Kadam$^{1}$, Shan Liu$^{2}$ and C.-C. Jay Kuo$^{1}$ }
\address{$^{1}$ Media Communications Lab, University of Southern California, Los Angeles, CA, USA \\ 
$^{2}$ Tencent Media Lab, Tencent America, Palo Alto, CA, USA}

\maketitle
\ninept 

\begin{abstract}

The PointHop method was recently proposed by Zhang {\em et al.} for 3D point cloud classification with unsupervised feature extraction. It has an extremely low training complexity while achieving state-of-the-art classification performance. In this work, we improve the PointHop method furthermore in two aspects: 1) reducing its model complexity in terms of the model parameter number and 2) ordering discriminant features automatically based on the cross-entropy criterion. The resulting method is called PointHop++. The first improvement is essential for wearable and mobile computing while the second improvement bridges statistics-based and optimization-based machine learning methodologies. With experiments conducted on the ModelNet40 benchmark dataset, we show that the PointHop++ method performs on par with deep neural network (DNN) solutions and surpasses other unsupervised feature extraction methods. 

\end{abstract}


\section{Introduction} 
\label{sec:intro}

Point cloud data processing find numerous applications such as computer-aided design (CAD) and AR/VR. It is however well known that the irregular and unordered distribution of points in the 3D space makes point cloud classification, segmentation and recognition very challenging. Although being successfully applied to 2D images \cite{krizhevsky2012imagenet, simonyan2014very, szegedy2015going, he2016deep}, deep learning techniques face several challenges in the context of 3D point cloud processing. To tackle with them, it is often to convert point clouds to other forms such as voxel grids, meshes and multi-view images. Afterwards, they can be processed by Convolutional Neural Network (CNN) methods \cite{maturana2015voxnet, su2015multi, zhou2018voxelnet, feng2018meshnet, feng2018gvcnn}. As compared with methods using raw point clouds as the input, conversion-based methods do have information loss. Besides, they demand additional memory and computation. Recently, we have seen a new trend that builds end-to-end deep networks to process point clouds directly \cite{qi2017pointnet, qi2017pointnet++, li2018pointcnn, wang2018dynamic} with the PointNet \cite{qi2017pointnet} as an example. 

Being inspired by recent work on feedforward-designed CNNs \cite{kuo2019interpretable}, the PointHop method was proposed in \cite{zhang2019pointhop} for point clouds classification. Its design was built upon the successive subspace learning (SSL) principle \cite{chen2020pixelhop}. PointHop consists of several PointHop units in cascade, and each of them comprises of neighborhood points search, quadrant-space-based feature representation, and dimension reduction. Attributes of a point are determined by the distribution of its neighboring points. The Saab transform \cite{kuo2019interpretable} is used to control the rapid increase in the attribute size. The local-to-global attributes of 3D point clouds can be obtained through an iterative process of one-hop information exchange. They are fed into a classifier to yield the final classification result. PointHop achieves classification accuracy similar to that of PointNet \cite{qi2017pointnet} yet demanding much less training and inference time. 

Here, we improve PointHop furthermore in two aspects: 1) reducing its model complexity in terms of the model parameters number; and 2) automatic selection of discriminant features based on the cross-entropy criterion. The resulting method is called PointHop++. The first improvement is essential for wearable and mobile computing \cite{iandola2016squeezenet, iandola2016firecaffe, qiu2016going} while the second improvement bridges statistics-based and optimization-based machine learning methodologies. With experiments conducted on the ModelNet40 dataset, we show that PointHop++ performs on par with CNN solutions and surpasses other unsupervised feature extraction methods. 

The rest of this paper is organized as follows. Background review is given in Sec. \ref{sec:background}. The PointHop++ method is detailed in Sec. \ref{sec:method}. Experimental results are shown in Sec. \ref{sec:experiment}. Finally, concluding remarks are given in Sec. \ref{sec:conclusion}. 

\section{Background Review} 
\label{sec:background}

\begin{figure*}[htp]
\centering
\includegraphics[width=17.8cm]{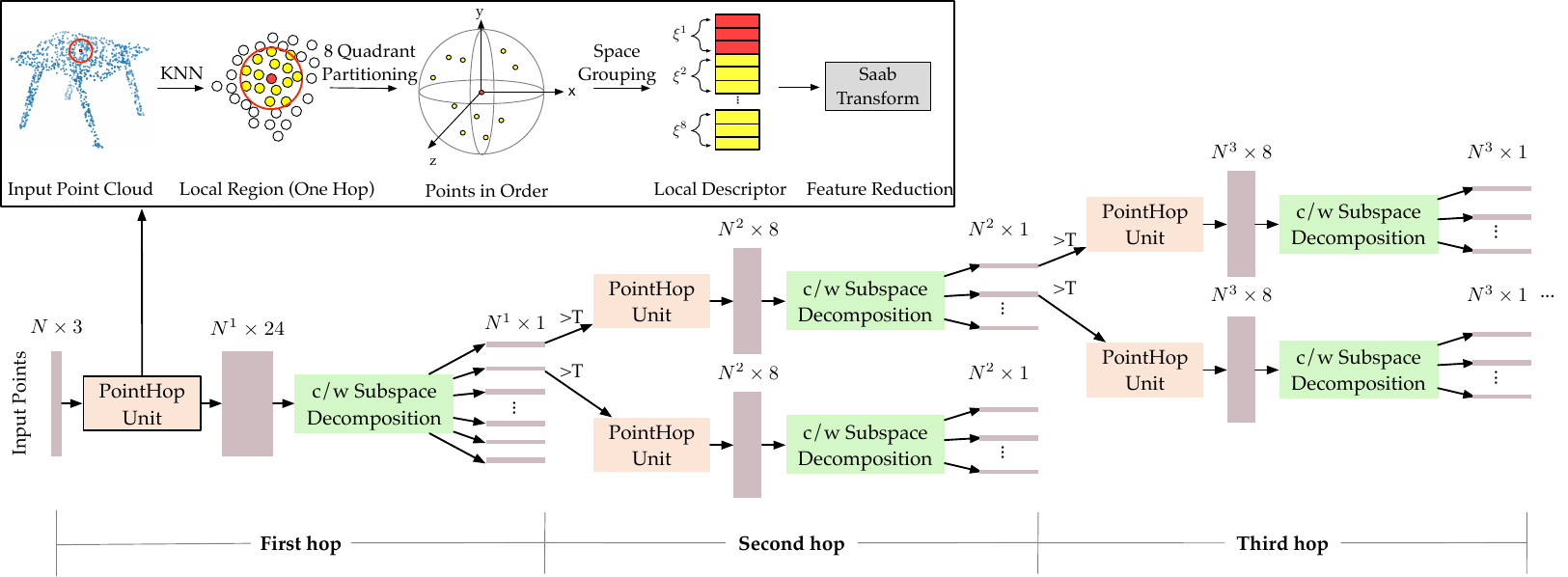}
\caption{Illustration of the PointHop++ method, where the upper-left enclosed subfigure shows the operation in the first PointHop unit, and $N$ and $N^i$ denote the number of points of the input and in the $n$th hop, respectively. Due to little correlation between channels, we can perform channel-wise (c/w) subspace decomposition to reduce the model size. A subspace with its energy larger than threshold $T$ proceeds to the next hop while others become leaf nodes of the feature tree in the current hop.}\label{fig:arch}
\end{figure*}

Deep-learning-based point cloud processing methods have several shortcomings {\em e.g.}, long training time, larger model sizes, use of expensive GPU resources and vulnerability to adversarial attacks. On top of them, they are mathematically intractable and difficult to interpret. Research has been performed to shed light on CNNs \cite{kuo2016understanding, kuo2017cnn, kuo2018data}. Kuo {\em et al.} proposed an unsupervised feature extraction method using an subspace approximation idea. Specifically, he and his co-authors introduced the Saak transform and the Saab transform in \cite{chen2018saak} and \cite{kuo2019interpretable}, respectively, and conducted them in multi-stages successively. The subspace approximation idea plays a role similar to the convolution layer of CNNs. Yet, no backpropagation (BP) is needed in Saak and Saab filters design. They are derived from statistics of pixels of input images without any label information. By following this line of thought, PointHop was proposed in \cite{zhang2019pointhop} for point cloud classification, where no BP is needed. 

There are two shortcomings of PointHop. First, it has a large spatial dimension and a small spectral dimension in the beginning of the pipeline. Each point has a small receptive field. As we move to further hops (or stages), the receptive field increases in size, and the system trades a larger spatial dimension for a higher spectral dimension. We use $n_t=n_a \times n_e$ to denote the tensor dimension at a certain hop, where $n_a$ and $n_e$ are spatial and spectral dimensions, respectively. Under the SSL framework, we need to conduct the principal component analysis (PCA) on input tensor space. That is, we compute the covariance matrix of vectorized tensors, which has a dimension of $n_t \times n_t$. Then, if we want to find $d$ principal components, the complexity is $O(d n_t^2 + d^3)$. Since $n_t > d$, the first term dominates. To make the learning model smaller, it is desired to lower the input tensor dimension so as to reduce the filter size. Second, the loss function minimization plays an important role in deep-learning-based methods. However, it was not incorporated in PointHop. To get a lightweight model and leverage the loss function for better performance, we present new ideas to improve PointHop. 

The current work has two major contributions. First, we show that the correlation between different spectral channels is very weak in Sec. \ref{sec:method}. Thus, we can decouple one joint spatial/spectral tensor of dimension $(n_a \times n_e)$ into $n_e$ spatial tensors of dimension $n_a$. Each of them is associated with a single spectral component. It is called the channel-wise (c/w) subspace decomposition. This idea helps reduce the model complexity of PointHop in its model parameters number and computational memory requirement. Second, through multiple decomposition stages, we obtain a one-dimensional (1D) feature at each leaf node of a feature tree. We use the cross-entropy loss function to rank features so that we can select a subset of discriminant features to train classifiers. This bridges statistics-based and optimization-based machine learning methodologies. 

\section{Proposed PointHop++ Method}
\label{sec:method}

An overview of the proposed PointHop++ method is illustrated in Fig. \ref{fig:arch}. A point cloud set, ${\bf P}$, which consists of $N$ points denoted by $p_n=(x_n, y_n, z_n)$, $1\leq n \leq N$, is taken as input to the feature learning system to obtain a powerful feature representation. After that, the linear least squares regression (LLSR) is conducted on the obtained features to output the 40D probability vector where the corresponding class labels come from. 

This section is organized as follows. The initial feature space construction is discussed in Sec. \ref{ssec:method_1}. The channel-wise subspace decomposition is presented in Sec. \ref{ssec:method_2}. Feature priority ordering is examined in Sec. \ref{ssec:method_3}. Finally, the PointHop++ method is detailed in Sec. \ref{ssec:method_4}. 

\subsection{Initial Feature Space Construction} 
\label{ssec:method_1}

Given a point cloud, $P=\{p_1, p_2, \cdots, p_N\}$, where ${p}_n \in \mathbb{R}^3$, $N$ is the size of the point set. To extract the local feature of each point ${p}_c \in P$, we follow the same design principle of the PointHop unit. The $k$ nearest neighbor points of point ${p}_c$ are retrieved to build a neighboring point set:
$$
{\mbox Neighborhood} (p_c) = \{p_{c_1}, p_{c_2}, \cdots, p_{c_k}\},
$$
including $p_c$ itself. The neighborhood set excluding $p_c$ is partitioned into eight quadrants according to their relative spatial coordinates. Then, the mean pooling is used to generate a $D$-dimensional attribute vector of each quadrant. Mathematically, we have the following mapping:
\begin{equation}
g: \underbrace{\mathbb{R}^{D} \times \cdots \mathbb{R}^{D}}_{\text{k}} 
\rightarrow \underbrace{\mathbb{R}^{D} \times \cdots \mathbb{R}^{D}}_{\text{8}},
\end{equation}

where $D=3$ for the first hop and $D=1$ for the remaining hops. The operation in the first PointHop unit is shown in the upper-left enclosed subfigure of Fig. \ref{fig:arch}. In words, the averaged attribute of all points in a quadrant is selected as the representative attribute of that quadrant. For the first hop, we use the spatial coordinates $p_n=(x_n, y_n, z_n)$ as the attributes. For the remaining hops, we use a one-dimensional (1D) spectral component as the attribute of retrieved points. This is possible since we apply the c/w subspace decomposition to the output from the previous hop. 

It is worthwhile to point out that, instead of using the max pooling as a symmetric function, we adopt the mean pooling as a symmetry function here. This is to ensure that the attributes of points are invariant under the permutation of points in the point cloud while the local structure is retained at the same time. Attributes of all eight quadrants are concatenated to become $ {\bf a} \in \mathbb{R}^{8D}$, which represents the attribute of selected point $p_c$ before c/w subspace decomposition. 

\subsection{Channel-Wise (C/W) Subspace Decomposition}
\label{ssec:method_2}

The Saab transform \cite{kuo2019interpretable} is a variant of the PCA \cite{wold1987principal} designed to overcome the sign confusion problem \cite{kuo2016understanding} when multiple PCA stages are in cascade. It is used as a dimension reduction tool in PointHop. All Saab transform coefficients are grouped together and used as the input to the next hop unit in PointHop. Here, we would like to prove that the Saab coefficients of different channels are weakly correlated. Then, we can decompose the Saab coefficient vector of dimension $8D$ into $8$ one-dimensional (1D) subspaces. Each 1D subspace represents a spatial-spectral localized representation of the point set. Besides its physical meaning, this representation demands less computation in the next hop. For ease of implementation, all components after the Saab transform are kept in PointHop++. 

To validate c/w subspace decomposition, we compute the correlation of Saab coefficients. The input to the Saab transform is
$$ 
A = [{\bf a}^1, \cdots, {\bf a}^N]^T \in \mathbb{R}^{N \times 8D},
$$ 
where ${\bf a}^n$ is the 8D attribute vector of point $p_n$, and the filter weight is
$$
W = [{\bf w}_1, {\bf w}_2, \cdots, {\bf w}_{8D}] \in \mathbb{R}^{8D \times 8D},
$$ 
where ${\bf w}_1 = \frac{1}{\sqrt{8D}}[1, 1, \cdots, 1]^T$ and others are eigenvectors of covariance matrix $A$ ranked by its associated eigenvalue $\lambda_i$ from the largest to the smallest. The output of the Saab transform is
$$
B = A \cdot W = [{\bf b}_1, \cdots, {\bf b}_{8D}], 
$$
where ${\bf b}_i \in \mathbb{R}^{N \times 1}$, $i=1, \cdots, 8D$. Hence, the correlation between Saab coefficients of different channels is
\begin{equation}
\begin{aligned}
Cor({\bf{b}_i},{\bf{b}_j})&= \frac{1}{N}(A \cdot {\bf w}_i)^T (A \cdot {\bf w}_j) 
= \frac{1}{N}(\lambda_i {\bf w}_i)^T (\lambda_j {\bf w}_j)\\
&=0,
\end{aligned}
\end{equation}
where $i \neq j$. The last equality comes from the orthogonality of eigenvectors in PCA analysis. This justifies the decomposition of a joint feature space into multiple uncorrelated 1D subspaces as
\begin{equation}
\mathbb{R}^{8D} \rightarrow \underbrace{\mathbb{R}^{1} \times \cdots 
\mathbb{R}^{1}}_{\text{8D}}.
\end{equation}
We should point out that, because of the special choice of the first filter weight ${\bf w}_1$, the above analysis is only an approximation. In practice, we observe very weak correlation between Saab coefficients (in the order of $10^{-4}$) as compared to the diagonal term (i.e. self-correlation).). 

\subsection{Channel Split Termination and Feature Priority Ordering}
\label{ssec:method_3}

We compute the energy of each subspace as
\begin{equation}
E_i = E_p \times \frac{\lambda_i}{\sum_{j=1}^{8D}{\lambda_j}},
\end{equation}
where $i=1, \cdots, 8D$ and $E_p$ is the energy of its parent node. If the energy of a node is less than a pre-set threshold, $T$, we terminate its further split and keep it as a leaf node of the feature tree at the current hop. Other nodes will proceed to the next hop. All leaf nodes are collected as the feature representation after the feature tree construction is completed. 

\begin{figure}[t]
\begin{minipage}[b]{0.49\linewidth}
  \centering
  \centerline{\includegraphics[width=4.3cm]{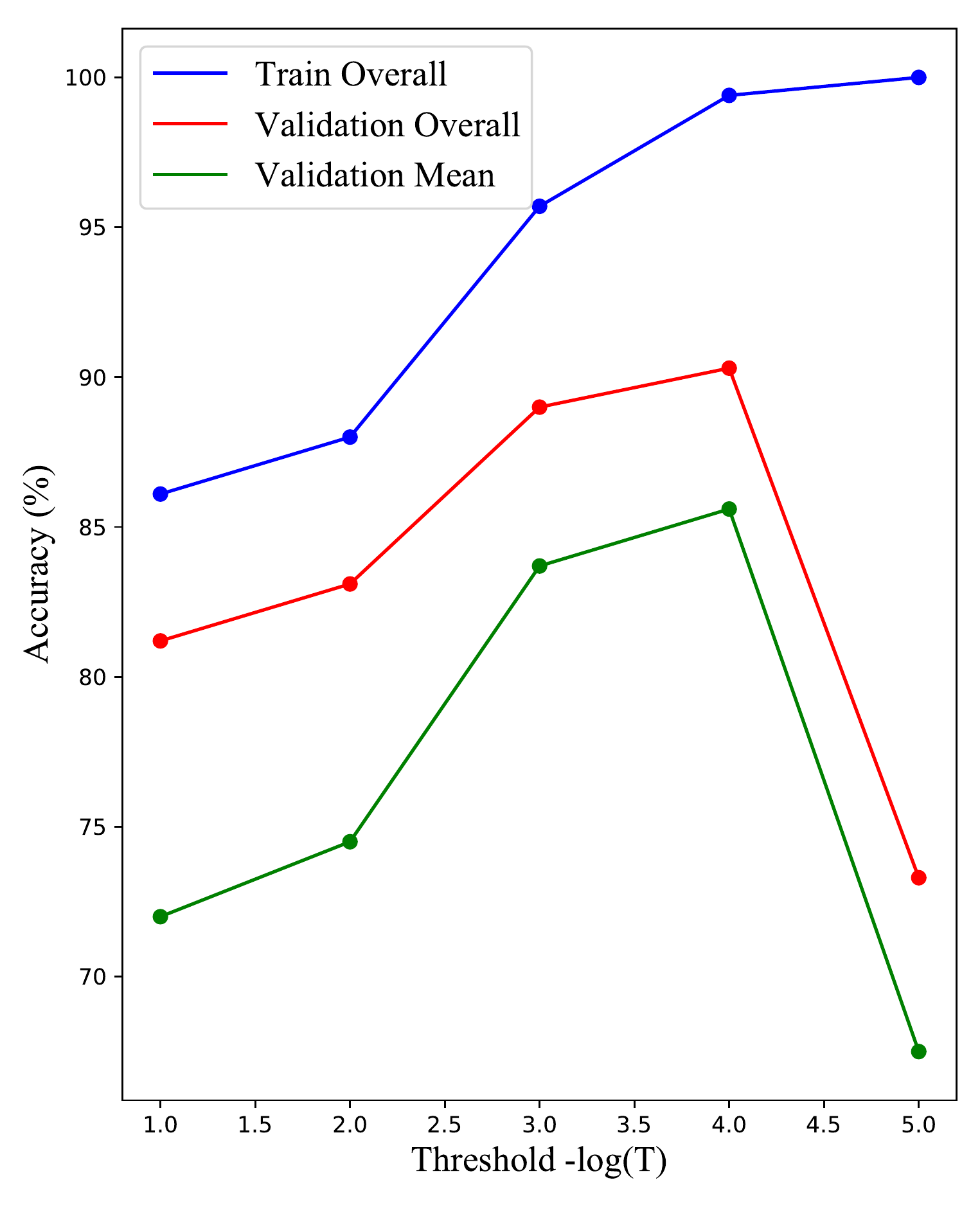}}
  \centerline{(a) Value of Threshold}\medskip
\end{minipage}
\begin{minipage}[b]{0.49\linewidth}
  \centering
  \centerline{\includegraphics[width=4.3cm]{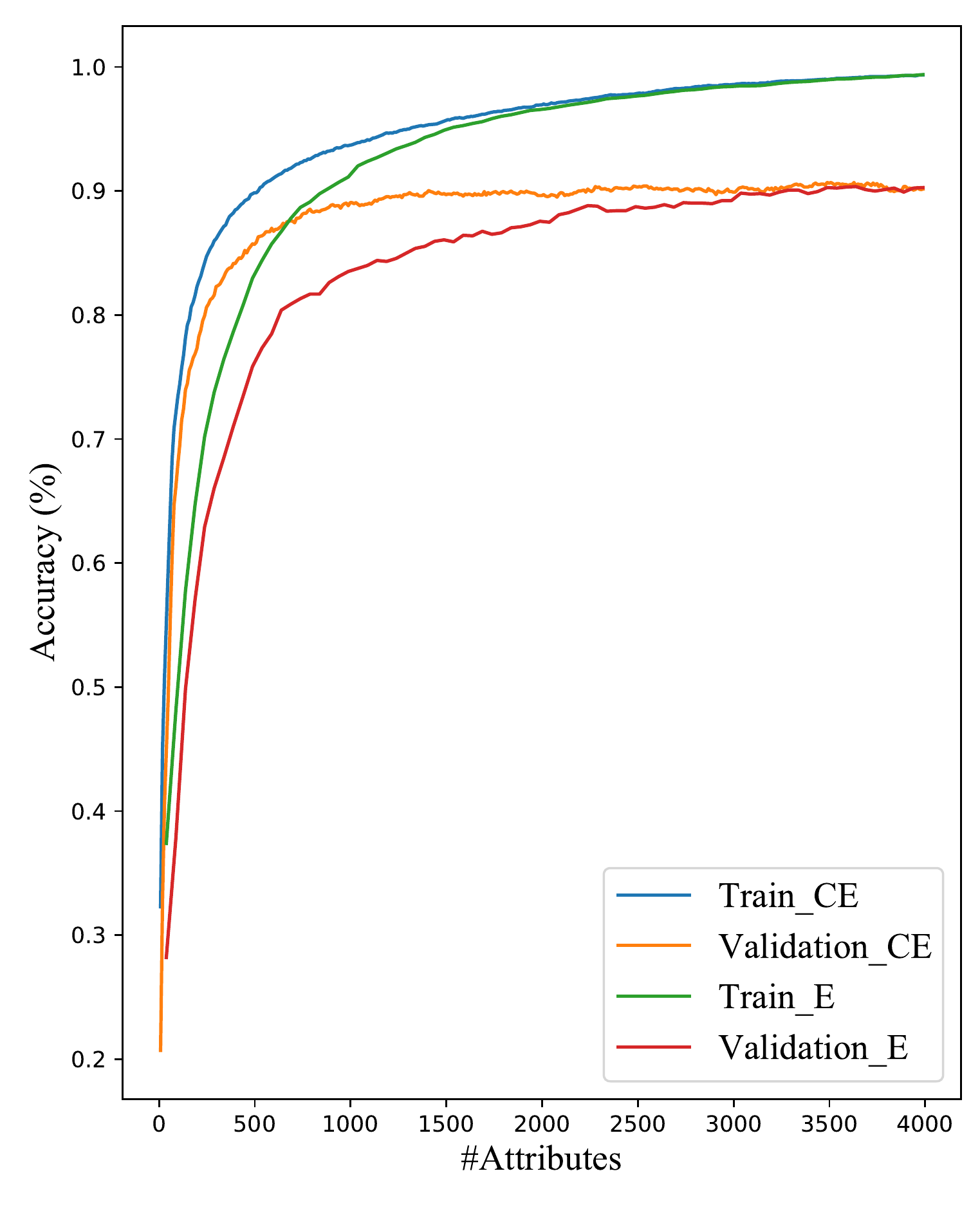}}
  \centerline{(b) Number of Ranked Features}\medskip
\end{minipage}
\caption{Illustration of the impact of (a) values of the energy threshold and (b) the number of cross-entropy-ranked (CE) or energy-ranked (E) features.} \label{fig:val}
\end{figure}

To determine threshold value $T$, the training and validation accuracy curves are plotted as a function of $T$ in Fig. \ref{fig:val} (a). We see that the training accuracy keeps increasing when $T$ decrease from 0.1 to 0.00001. Yet, the overall validation accuracy increases to the maximum value of 90.3\% at $T=0.0001$. After that, the validation accuracy decreases. Thus, we choose $T=0.0001$. 

Once the feature tree is constructed, it is desired to order features based on their discriminant power and select them accordingly to avoid overfit. A feature is more discriminant if its cross entropy is lower. The cross entropy can be computed for each feature at a leaf node. We follow the same process as described in \cite{kuo2019interpretable}. That is, a clustering algorithm \cite{wagstaff2001constrained} is adopted to partition the 1D subspace into $J$ intervals. Then, the majority vote is used to predict the label for each interval. Based on the groundtruth labels, the probability that each sample belongs to a class can be obtained. Mathematically, we have
\begin{equation}
L =\sum_{j=1}^{J}L_j, \quad L_j =-\sum_{c=1}^{M}y_{j,c}log(p_{j,c}), 
\end{equation}
where $M$ is the class number, $y_{j,c}$ is binary indicator to show whether sample $j$ is correctly classified, and $p_{j,c}$ is the probability that sample $j$ belongs to class $c$.

We compare training and validation accuracy curves using features that are ranked by the cross-entropy values and the energy values, respectively, in Fig. \ref{fig:val} (b), where the x-axis indicates the total number of top-ranked features. We see that overfitting is improved by both methods. The cross-entropy-ranked method performs better when the total feature number is smaller. 

\subsection{Summary of PointHop++ Method}
\label{ssec:method_4}

The tree-structured feature construction process at each hop can be summarized as follows.
\begin{itemize}
\item Use the knn algorithm to retrieve neighbor points;
\item Use the decoupled attribute to perform the Saab transform;
\item If the energy of a node is greater than a pre-set threshold, perform the c/w subspace decomposition and obtain decoupled attributes as the input to the next hop. 
\end{itemize}
The above process is repeated until the last hop is reached. Once the feature tree construction is completed, each leaf node contains a scalar feature. These features are ranked according to their energy and cross entropy. Finally, the LLSR is adopted as the classifier. 

\section{Experiments}
\label{sec:experiment}

Experiments are conducted on the ModelNet40 dataset \cite{wu20153d}, which contains 40 object classes. 1024 points are sampled randomly from the original point cloud set as the input to PointHop++. The depth of the feature tree is set to four hops. The farthest point sampling \cite{eldar1997farthest} is used to downsample points from one hop to the next to increase the receptive field and speed up the computation. 

\begin{table}[h]
\centering
\renewcommand\arraystretch{1.1}
\begin{tabular}{c|c|c|c} \hline
& \multirow{2}*{Method} & \multicolumn{2}{c}{Accuracy (\%)} \\ \cline{3-4}
& & class-avg & overall  \\ \hline
\multirow{4}*{Supervised} & PointNet \cite{qi2017pointnet}  & 86.2 & 89.2 \\
& PointNet++ \cite{qi2017pointnet++} & - & 90.7 \\
& PointCNN \cite{li2018pointcnn} & 88.1 & \bf{92.2} \\
& DGCNN \cite{wang2018dynamic} & \bf{90.2} & \bf{92.2} \\ \hline
\multirow{4}*{Unsupervised} & LFD-GAN \cite{achlioptas2017representation} & - & 85.7 \\
& FoldingNet \cite{yang2018foldingnet} & - & 88.4 \\
& PointHop \cite{zhang2019pointhop} & 84.4 & 89.1 \\ \cline{2-4}
& PointHop++ (baseline) & 85.6 & 90.3 \\ 
& PointHop++ (FS) & 86.5 & 90.8 \\ 
& PointHop++ (FS+ES) & \bf{87} & \bf{91.1} \\ \hline
\end{tabular}
\caption{Comparison of classification results on ModelNet40, where the class-Avg accuracy is the mean of the per-class accuracy, and FS and ES mean ``feature selection" and ``ensemble", respectively.}
\label{tab:comp_acc}
\end{table}

\begin{table}[h]
\centering
\renewcommand\arraystretch{1.1}
\newcommand{\tabincell}[2]{\begin{tabular}{@{}#1@{}}#2\end{tabular}}
\setlength{\tabcolsep}{0.8mm}{
\begin{tabular}{c|cc|ccc} \hline
\multirow{2}*{Method} & \multicolumn{2}{c|}{Time} & \multicolumn{3}{c}{Parameter No. (MB)} \\ \cline{2-6}
& Training & Inference & Filter & Classifier & Total \\ \hline
PointNet \cite{qi2017pointnet} & 7 & 10 & - & - & 3.48 \\ 
PointNet++ \cite{qi2017pointnet++} & 7 & 14 & - & - & 1.48 \\ 
DGCNN \cite{wang2018dynamic} & 21 & 154 & - & - & 1.84 \\ \hline
PointHop \cite{zhang2019pointhop} & 0.33 & 108 & 0.037 & - & - \\ 
PointHop++ & 0.42 & 97 & 0.009 & 0.15 & 0.159 \\ \hline
\end{tabular}}
\caption{Comparison of time and model complexity, where the training and inference time units are in hour and ms, respectively.} 
\label{tab:comp_complexity}
\end{table}

Classification accuracy of different methods are compared in Table \ref{tab:comp_acc}. PointHop++ (baseline), which has an energy threshold 0.0001 without feature selection or ensembles, gives 90.3\% overall accuracy and 85.6\% class-avg accuracy. By incorporating the feature selection tool as discussed in Sec. \ref{ssec:method_3}, PointHop++ (FS) improves the overall and class-avg accuracy results by 0.5\% and 0.9\%, respectively.  Furthermore, we rotate point clouds by 45 degrees and conduct LLSR to get a 40D feature for eight times. Then, these features are concatenated and fed into another LLSR. The ensemble method has an overall accuracy of 91.1\% and a class-avg accuracy of 87\%. PointHop++ method achieves the best performance among unsupervised feature extraction methods. It outperforms PointHop \cite{zhang2019pointhop} by 2\% in overall accuracy. As compared with deep networks, PointHop++ outperforms PointNet \cite{qi2017pointnet} and PointNet++ \cite{qi2017pointnet++}. It has a gap of 1.1\% against PointCNN \cite{li2018pointcnn} and DGCNN \cite{wang2018dynamic}. 

Comparison of time complexity and model sizes of different methods is given in Table \ref{tab:comp_complexity}. Four deep networks were trained on a single GeForce GTX TITAN X GPU. It took at least 7 hours to train a 1,024 point cloud model while PointHop++ only took 25 minutes on a Intel(R)Xeon(R) CPU. As to inference time of every sample, both PointHop and PointHop++ took about 100 ms while DGCNN took 163 ms. The number of model parameters are also computed to show space complexity. The Saab filter size of PointHop++ is 4X less than that of PointHop. The total model parameters of PointHop++ is 20X less than that PointNet \cite{qi2017pointnet} and 10X less than DGCNN \cite{wang2018dynamic} \cite{li2018pointcnn}. 

\begin{figure}[t]
\centering
\includegraphics[width=6.5cm]{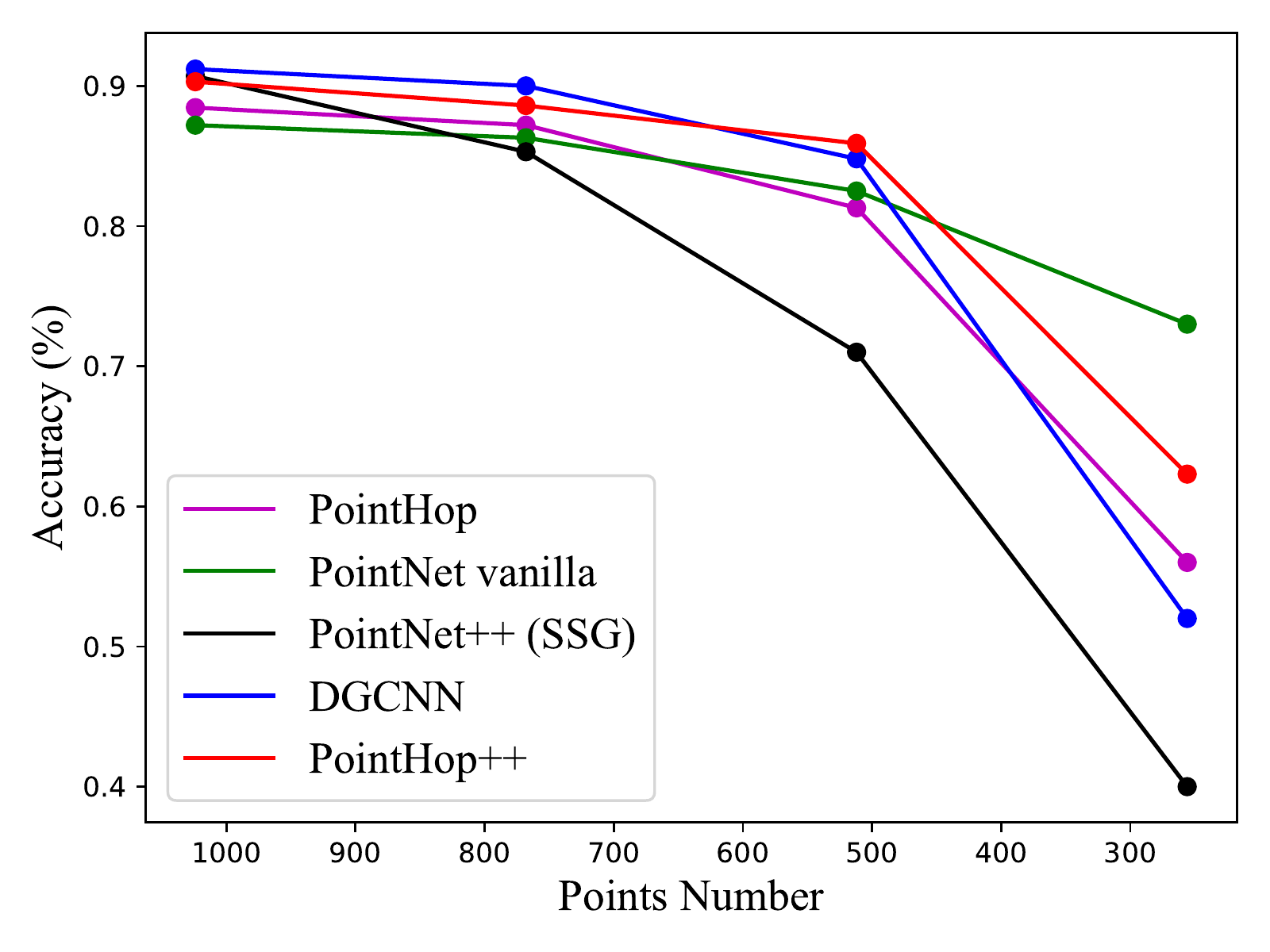}
\caption{Robustness against different sampling densities of the test model.}
\label{fig:robust}
\end{figure}

We compare the robustness of different models against sampling density variation in Fig. \ref{fig:robust}. All models are trained on 1,024 point cloud model. The test model are randomly sampled with 768, 512, and 256 points, respectively. We see that PointHop++ are more robust than PointHop \cite{zhang2019pointhop}, PointNet++ (SSG) \cite{qi2017pointnet++} and DGCNN \cite{wang2018dynamic} under mismatched sampling densities.  

\begin{figure}[h]
\begin{minipage}[b]{0.48\linewidth}
  \centering
  \centerline{\includegraphics[width=4.6cm]{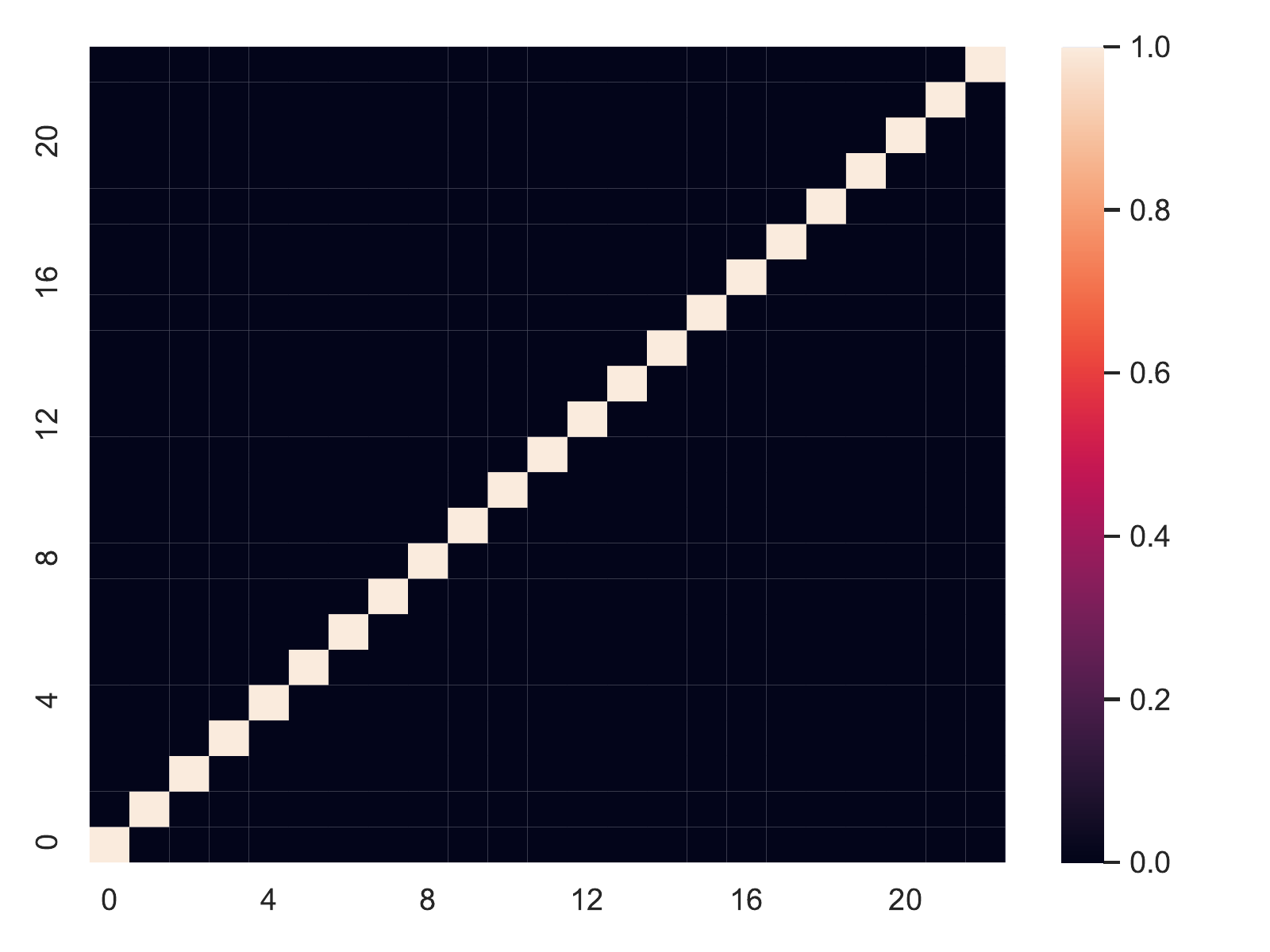}}
  \centerline{(a) correlation matrix}\medskip
\end{minipage}
\begin{minipage}[b]{0.48\linewidth}
  \centering
  \centerline{\includegraphics[width=4.4cm]{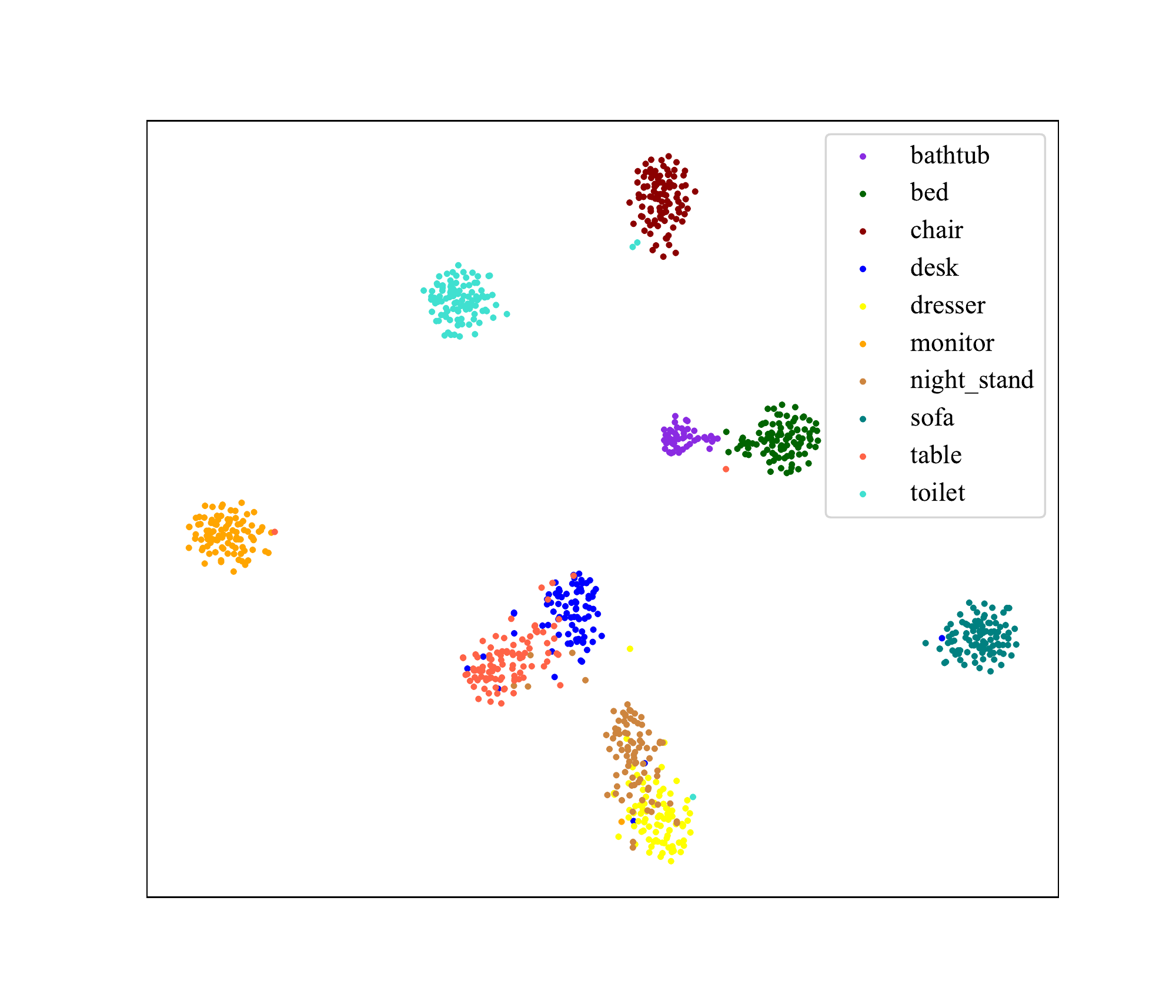}}
  \centerline{(a) feature clustering}\medskip
\end{minipage}
\caption{Visualization of (a) the correlation matrix at the first hop and (b) feature clustering in the T-SNE plot.}
\label{fig:vis}
\end{figure}

Finally, we show the correlation matrix of AC components at the first hop in Fig. \ref{fig:vis}. It verifies the claim that different AC components are uncorrelated. Furthermore, we visualize the feature distribution with the T-SNE plot, where the dimension is reduced to 2D. We visualize the features of the 10 object classes from ModelNet10 \cite{wu20153d}, which is a subset of ModelNet40 \cite{wu20153d}. We see that most features of the same category are clustered together, which demonstrates the discriminant power of features selected by PointHop++. 

\section{Conclusion}
\label{sec:conclusion}

A tree-structured unsupervised feature learning system was proposed in this work, where one scalar feature is associated with each leaf node and features are ordered based on their discriminant power. The resulting PointHop++ method achieves state-of-the-art classification performance while demanding a significantly small learning model which is ideal for mobile computing. 

\newpage
\bibliographystyle{IEEEbib}
\bibliography{strings,refs}

\begin{thebibliography}{10}

\bibitem{krizhevsky2012imagenet}
Alex Krizhevsky, Ilya Sutskever, and Geoffrey~E Hinton,
\newblock ``Imagenet classification with deep convolutional neural networks,''
\newblock in {\em Advances in neural information processing systems}, 2012, pp.
  1097--1105.

\bibitem{simonyan2014very}
Karen Simonyan and Andrew Zisserman,
\newblock ``Very deep convolutional networks for large-scale image
  recognition,''
\newblock {\em arXiv preprint arXiv:1409.1556}, 2014.

\bibitem{szegedy2015going}
Christian Szegedy, Wei Liu, Yangqing Jia, Pierre Sermanet, Scott Reed, Dragomir
  Anguelov, Dumitru Erhan, Vincent Vanhoucke, and Andrew Rabinovich,
\newblock ``Going deeper with convolutions,''
\newblock in {\em Proceedings of the IEEE conference on computer vision and
  pattern recognition}, 2015, pp. 1--9.

\bibitem{he2016deep}
Kaiming He, Xiangyu Zhang, Shaoqing Ren, and Jian Sun,
\newblock ``Deep residual learning for image recognition,''
\newblock in {\em Proceedings of the IEEE conference on computer vision and
  pattern recognition}, 2016, pp. 770--778.

\bibitem{maturana2015voxnet}
Daniel Maturana and Sebastian Scherer,
\newblock ``Voxnet: A 3d convolutional neural network for real-time object
  recognition,''
\newblock in {\em 2015 IEEE/RSJ International Conference on Intelligent Robots
  and Systems (IROS)}. IEEE, 2015, pp. 922--928.

\bibitem{su2015multi}
Hang Su, Subhransu Maji, Evangelos Kalogerakis, and Erik Learned-Miller,
\newblock ``Multi-view convolutional neural networks for 3d shape
  recognition,''
\newblock in {\em Proceedings of the IEEE international conference on computer
  vision}, 2015, pp. 945--953.

\bibitem{zhou2018voxelnet}
Yin Zhou and Oncel Tuzel,
\newblock ``Voxelnet: End-to-end learning for point cloud based 3d object
  detection,''
\newblock in {\em Proceedings of the IEEE Conference on Computer Vision and
  Pattern Recognition}, 2018, pp. 4490--4499.

\bibitem{feng2018meshnet}
Yutong Feng, Yifan Feng, Haoxuan You, Xibin Zhao, and Yue Gao,
\newblock ``Meshnet: Mesh neural network for 3d shape representation,''
\newblock {\em arXiv preprint arXiv:1811.11424}, 2018.

\bibitem{feng2018gvcnn}
Yifan Feng, Zizhao Zhang, Xibin Zhao, Rongrong Ji, and Yue Gao,
\newblock ``Gvcnn: Group-view convolutional neural networks for 3d shape
  recognition,''
\newblock in {\em Proceedings of the IEEE Conference on Computer Vision and
  Pattern Recognition}, 2018, pp. 264--272.

\bibitem{qi2017pointnet}
Charles~R Qi, Hao Su, Kaichun Mo, and Leonidas~J Guibas,
\newblock ``Pointnet: Deep learning on point sets for 3d classification and
  segmentation,''
\newblock in {\em Proceedings of the IEEE Conference on Computer Vision and
  Pattern Recognition}, 2017, pp. 652--660.

\bibitem{qi2017pointnet++}
Charles~Ruizhongtai Qi, Li~Yi, Hao Su, and Leonidas~J Guibas,
\newblock ``Pointnet++: Deep hierarchical feature learning on point sets in a
  metric space,''
\newblock in {\em Advances in Neural Information Processing Systems}, 2017, pp.
  5099--5108.

\bibitem{li2018pointcnn}
Yangyan Li, Rui Bu, Mingchao Sun, Wei Wu, Xinhan Di, and Baoquan Chen,
\newblock ``Pointcnn: Convolution on x-transformed points,''
\newblock in {\em Advances in Neural Information Processing Systems}, 2018, pp.
  820--830.

\bibitem{wang2018dynamic}
Yue Wang, Yongbin Sun, Ziwei Liu, Sanjay~E Sarma, Michael~M Bronstein, and
  Justin~M Solomon,
\newblock ``Dynamic graph cnn for learning on point clouds,''
\newblock {\em arXiv preprint arXiv:1801.07829}, 2018.

\bibitem{kuo2019interpretable}
C-C~Jay Kuo, Min Zhang, Siyang Li, Jiali Duan, and Yueru Chen,
\newblock ``Interpretable convolutional neural networks via feedforward
  design,''
\newblock {\em Journal of Visual Communication and Image Representation}, vol.
  60, pp. 346--359, 2019.

\bibitem{zhang2019pointhop}
Min Zhang, Haoxuan You, Pranav Kadam, Shan Liu, and C-C~Jay Kuo,
\newblock ``Pointhop: An explainable machine learning method for point cloud
  classification,''
\newblock {\em arXiv preprint arXiv:1907.12766}, 2019.

\bibitem{chen2020pixelhop}
Yueru Chen and C-C~Jay Kuo,
\newblock ``Pixelhop: A successive subspace learning (ssl) method for object
  recognition,''
\newblock {\em Journal of Visual Communication and Image Representation}, p.
  102749, 2020.

\bibitem{iandola2016squeezenet}
Forrest~N Iandola, Song Han, Matthew~W Moskewicz, Khalid Ashraf, William~J
  Dally, and Kurt Keutzer,
\newblock ``Squeezenet: Alexnet-level accuracy with 50x fewer parameters and<
  0.5 mb model size,''
\newblock {\em arXiv preprint arXiv:1602.07360}, 2016.

\bibitem{iandola2016firecaffe}
Forrest~N Iandola, Matthew~W Moskewicz, Khalid Ashraf, and Kurt Keutzer,
\newblock ``Firecaffe: near-linear acceleration of deep neural network training
  on compute clusters,''
\newblock in {\em Proceedings of the IEEE Conference on Computer Vision and
  Pattern Recognition}, 2016, pp. 2592--2600.

\bibitem{qiu2016going}
Jiantao Qiu, Jie Wang, Song Yao, Kaiyuan Guo, Boxun Li, Erjin Zhou, Jincheng
  Yu, Tianqi Tang, Ningyi Xu, Sen Song, et~al.,
\newblock ``Going deeper with embedded fpga platform for convolutional neural
  network,''
\newblock in {\em Proceedings of the 2016 ACM/SIGDA International Symposium on
  Field-Programmable Gate Arrays}, 2016, pp. 26--35.

\bibitem{kuo2016understanding}
C-C~Jay Kuo,
\newblock ``Understanding convolutional neural networks with a mathematical
  model,''
\newblock {\em Journal of Visual Communication and Image Representation}, vol.
  41, pp. 406--413, 2016.

\bibitem{kuo2017cnn}
C-C~Jay Kuo,
\newblock ``The cnn as a guided multilayer recos transform [lecture notes],''
\newblock {\em IEEE signal processing magazine}, vol. 34, no. 3, pp. 81--89,
  2017.

\bibitem{kuo2018data}
C-C~Jay Kuo and Yueru Chen,
\newblock ``On data-driven saak transform,''
\newblock {\em Journal of Visual Communication and Image Representation}, vol.
  50, pp. 237--246, 2018.

\bibitem{chen2018saak}
Yueru Chen, Zhuwei Xu, Shanshan Cai, Yujian Lang, and C-C~Jay Kuo,
\newblock ``A saak transform approach to efficient, scalable and robust
  handwritten digits recognition,''
\newblock in {\em 2018 Picture Coding Symposium (PCS)}. IEEE, 2018, pp.
  174--178.

\bibitem{wold1987principal}
Svante Wold, Kim Esbensen, and Paul Geladi,
\newblock ``Principal component analysis,''
\newblock {\em Chemometrics and intelligent laboratory systems}, vol. 2, no.
  1-3, pp. 37--52, 1987.

\bibitem{wagstaff2001constrained}
Kiri Wagstaff, Claire Cardie, Seth Rogers, Stefan Schr{\"o}dl, et~al.,
\newblock ``Constrained k-means clustering with background knowledge,''
\newblock in {\em Icml}, 2001, vol.~1, pp. 577--584.

\bibitem{wu20153d}
Zhirong Wu, Shuran Song, Aditya Khosla, Fisher Yu, Linguang Zhang, Xiaoou Tang,
  and Jianxiong Xiao,
\newblock ``3d shapenets: A deep representation for volumetric shapes,''
\newblock in {\em Proceedings of the IEEE conference on computer vision and
  pattern recognition}, 2015, pp. 1912--1920.

\bibitem{eldar1997farthest}
Yuval Eldar, Michael Lindenbaum, Moshe Porat, and Yehoshua~Y Zeevi,
\newblock ``The farthest point strategy for progressive image sampling,''
\newblock {\em IEEE Transactions on Image Processing}, vol. 6, no. 9, pp.
  1305--1315, 1997.

\bibitem{achlioptas2017representation}
Panos Achlioptas, Olga Diamanti, Ioannis Mitliagkas, and Leonidas Guibas,
\newblock ``Representation learning and adversarial generation of 3d point
  clouds,''
\newblock {\em arXiv preprint arXiv:1707.02392}, vol. 2, no. 3, pp. 4, 2017.

\bibitem{yang2018foldingnet}
Yaoqing Yang, Chen Feng, Yiru Shen, and Dong Tian,
\newblock ``Foldingnet: Point cloud auto-encoder via deep grid deformation,''
\newblock in {\em Proceedings of the IEEE Conference on Computer Vision and
  Pattern Recognition}, 2018, pp. 206--215.

\end{thebibliography}

\end{document}